\documentclass{winnower}
\usepackage{multirow}
\usepackage{algorithm, algorithmic}
\usepackage{amsthm, mathtools}
\usepackage{wrapfig, ragged2e}
\usepackage{xcolor, minibox}

\newcommand{\R}{\mathbb{R}}

\newcommand{\pf}[1]{\nabla f (#1)}

\newcommand{\pfik}[1]{\nabla f_{i_k}(#1)}
\newcommand{\pfi}[1]{\nabla f_{i}(#1)}
\newcommand{\pfr}[2]{\nabla f_{#1}(#2)}
\newcommand{\pFi}[1]{\nabla F_i (#1)}
\newcommand{\pFr}[2]{\nabla F_{#1}(#2)}

\newcommand{\norm}[1]{\lVert#1\rVert}
\newcommand{\normbig}[1]{\Big\lVert#1\Big\rVert}

\newcommand{\innr}[1]{\langle#1\rangle}
\newcommand{\innrbig}[1]{\big\langle#1\big\rangle}

\newcommand{\Eik}[1]{\mathbb{E}_{i_k} \big[#1\big]}
\newcommand{\EikBig}[1]{\mathbb{E}_{i_k} \Big[#1\Big]}
\newcommand{\Er}[2]{\mathbb{E}_{#1} \big[#2\big]}
\newcommand{\ErBig}[2]{\mathbb{E}_{#1} \Big[#2\Big]}
\newcommand{\E}[1]{\mathbb{E} \big[#1\big]}

\newcommand{\mleq}[1]{\overset{\mathclap{(#1)}}{\leq}}
\newcommand{\meq}[1]{\overset{\mathclap{(#1)}}{=}}

\newcommand{\mar}[1]{(#1)}
\newcommand{\xs}{x^{\star}}

\DeclareMathOperator*{\prox}{prox}

\newtheorem{sconvex}{Assumption}

\newtheorem{siam_nsc}{Theorem}

\newtheorem{siam_ns}{Corollary}
\newtheorem{three_point}{Lemma}
\newtheorem{sc_prox}[three_point]{Lemma}
\newtheorem{moreau_envelope}[three_point]{Lemma}
\newtheorem{variance_bound}[three_point]{Lemma}

\begin{document}

\title{Direct Acceleration of SAGA using Sampled Negative Momentum}

\author{Kaiwen Zhou\footnote{Department of Computer Science and Engineering, The Chinese University of Hong Kong, Sha Tin, N.T., Hong Kong SAR; e-mail: \texttt{kwzhou@cse.cuhk.edu.hk}}}

\pagestyle{plain}

\date{}

\maketitle

\centering {\color{red}\minibox[frame]{The AISTATS version of this paper is too pessimistic about the memory overhead of the proposed \\ method. We would like to specially thank Hadrien Hendrikx from INRIA for his suggestions on \\ reducing the memory complexity during AISTATS 2019.}}

\justifying

\begin{abstract}
Variance reduction is a simple and effective technique that accelerates convex (or non-convex) stochastic optimization. Among existing variance reduction methods, SVRG and SAGA adopt unbiased gradient estimators and are the most popular variance reduction methods in recent years. Although various accelerated variants of SVRG (e.g., Katyusha and Acc-Prox-SVRG) have been proposed, the direct acceleration of SAGA still remains unknown. In this paper, we propose a directly accelerated variant of SAGA using a novel Sampled Negative Momentum (SSNM), which achieves the best known oracle complexity for strongly convex problems (with known strong convexity parameter). Consequently, our work fills the void of directly accelerated SAGA.
\end{abstract}

\section{Introduction}

In this paper\footnote{\color{red} \texttt{v2} fixed a mistake in proving Theorem~\ref{thm:strongly_convex} and added some extensions and insights. \texttt{v3} polished writing. \texttt{v4} discussed some implementation concerns.}, we consider optimizing the following composite finite-sum problem, which arises frequently in machine learning and statistics such as supervised learning and regularized empirical risk minimization (ERM):
\begin{equation}
\label{prob_def}
\min_{x\in \mathbb{R}^d} \left\{F(x) \triangleq f(x) + h(x)\right\},
\end{equation}
where $f(x)\!=\!\frac{1}{n}\!\sum_{i=1}^{n}f_i(x)$ is an average of $n$ smooth and convex function $f_i(x)$, and $h(x)$ is a simple and convex (but possibly non-differentiable) function. Here, we also define $F_i(x) = f_i(x) + h(x)$ with $\pFi{x} = \pfi{x} + \partial h(x)$ and $\partial h(x)$ denotes a sub-gradient of $h(\cdot)$ at $x$, which will be used in the paper.

We focus on achieving a highly accurate solution for Problem~\eqref{prob_def}, although for practical optimization tasks, such as supervised learning, low empirical risk may result in a high generalization error. In this paper, we treat Problem~\eqref{prob_def} as a pure optimization problem.

\begin{table*}[t]
	\centering
	\caption{Comparison of some accelerated variants of SVRG and SAGA. Here, we regard using reductions or proximal point variants as ``Indirect'' acceleration.}
	\vspace{10pt}
	\begin{tabular}{|l|c|c|}
		\hline
		&Indirect  & Direct \\
		\hline
		\rule{0pt}{11pt} 
		SVRG (or Prox-SVRG)  &\multirow{2}{*}{APPA \& Catalyst}       &Katyusha \& MiG\\
		\cline{1-1}\cline{3-3}
		\rule{0pt}{10pt}
		\multirow{2}{*}{SAGA}   &  &\multirow{2}{*}{\textbf{SSNM}}       \\
		\cline{2-2}
		&\rule{0pt}{10pt} Point-SAGA & \\
		\hline
	\end{tabular}
	\label{table:acc_sum}
\end{table*}

When $F(\cdot)$ in Problem~(\ref{prob_def}) is strongly convex, traditional analysis shows that gradient descent (GD) yields a fast linear convergence rate but with a high per-iteration cost, and thus may not be suitable for problems with a very large $n$. As an alternative for large-scale problems, SGD~\citep{robbins:sgd} uses only one or a mini-batch of gradients in each iteration, and thus enjoys a significantly lower per-iteration complexity than GD. However, due to the undiminished variance of the gradient estimator, vanilla SGD is shown to yield only a sub-linear convergence rate. Recently, stochastic variance reduced methods (e.g., SAG~\citep{roux:sag}, SVRG~\citep{johnson:svrg}, SAGA~\citep{defazio:saga}, and their proximal variants, such as~\citep{schmidt:sag}, ~\citep{xiao:prox-svrg} and~\citep{koneeny:mini}) were proposed to solve Problem~\eqref{prob_def}. All these methods are equipped with various variance reduction techniques, which help them achieve low per-iteration complexities comparable with SGD and at the same time maintain a faster linear convergence rate than GD (including accelerated GD). In terms of oracle complexity\footnote{Oracle complexity in this paper, denoted by $\mathcal{O}(\cdot)$, is the number of calls to Incremental First-order Oracle (IFO) + Proximal operator Oracle (PO).}, these methods all achieve an $\mathcal{O}\!\left((n\!+\!\kappa)\log({1}/{\epsilon})\right)$ complexity\footnote{We denote $\kappa \triangleq \frac{L}{\mu}$ throughout the paper, which is known as the condition number of an $L$-smooth and $\mu$-strongly convex function.}, as compared with $\mathcal{O}(n\sqrt{\kappa}\log({1}/{\epsilon}))$ for accelerated deterministic methods (e.g., Nesterov's accelerated gradient descent~\citep{nesterov:co}).

Inspired by the acceleration technique proposed in Nesterov's accelerated gradient descent~\citep{nesterov:co}, accelerated variants of stochastic variance reduced methods have been proposed in recent years, such as Acc-Prox-SVRG~\citep{nitanda:svrg}, APCG~\citep{lin:APCG}, APPA~\citep{roy:appa}, Catalyst~\citep{lin:vrsg}, SPDC~\citep{zhang:spdc} and Katyusha~\citep{zhu:Katyusha}. Among these algorithms, APPA and Catalyst achieve acceleration by using some carefully designed reduction techniques, which, however, result in additional log factors in their overall oracle complexities. Katyusha, as the first directly accelerated variant of SVRG, introduced the idea of negative momentum (or Katyusha momentum): regarding the gradient estimator of SVRG
\[
\widetilde{\nabla} = \pfi{x} - \pfi{\tilde{x}} + \pf{\tilde{x}},
\] the negative momentum is a $(\tilde{x} - x)$ offset added (with decay) to each update in this epoch. One can interpreted it as the momentum provided by a previously randomly computed point. Then, by combining it with Nesterov's momentum,  Katyusha yields the best known\footnote{According to~\citep{arjevani2017limitations}, this rate can only be attained when $\mu$ is known. Without knowing $\mu$, the best known rate is $\mathcal{O}\!\left((n\!+\!\kappa)\log({1}/{\epsilon})\right)$ achieved by~\citep{lei2017less} and~\citep{xu2017adaptive}. We assume $\mu$ is known throughout the paper.} oracle complexity $\mathcal{O}((n + \sqrt{\kappa n})\log(1/\epsilon))$ for strongly convex problems. More recent work~\citep{kw:MiG} shows that adding only negative momentum to SVRG is enough to achieve the best known oracle complexity for strongly convex problems, which results in a simple and scalable algorithm called MiG. 

Although a considerable amount of work has been done for accelerating SVRG,  another popular stochastic variance reduced method, SAGA, does not have a directly accelerated variant until recently. Accelerating frameworks such as APPA or Catalyst can be used to accelerate SAGA, but the reduction techniques proposed in these works are always difficult to implement and may also result in additional log factors in the overall oracle complexity. A notable variant of SAGA is Point-SAGA~\citep{defazio:sagab}. Point-SAGA requires the proximal operator oracle of each $F_i(\cdot)$ and with the help of that, it can adopt a much larger learning rate than SAGA, which results in the accelerated complexity $\mathcal{O}((n\!+\!\sqrt{\kappa n})\log(1/\epsilon))$. Some accelerated variants of SVRG and SAGA are summarized in Table~\ref{table:acc_sum}. However, the proximal operator of each $F_i(\cdot)$ may not be efficiently computed in practice. Even for logistic regression, we need to run an individual loop (Newton's method) for its proximal operator oracle. Therefore, a directly accelerated variant of SAGA is of real interests.

Following the idea of adding only negative momentum to SVRG~\citep{kw:MiG}, we consider adding negative momentum to SAGA. However, unlike SVRG, which keeps a constant snapshot in each inner loop, the ``snapshot'' of SAGA is a table of points, each corresponding to the position that the component function gradient $\pfi{\cdot}$ was lastly evaluated. Thus, it is non-trivial to directly accelerate SAGA. In this paper, we propose a novel \textit{Sampled Negative Momentum} for SAGA. We further show that adding such a momentum has the same acceleration effect as adding negative momentum to SVRG. 

Our contributions are summarized below:
\begin{itemize}
	\item We propose a directly accelerated variant of SAGA. The acceleration technique is a combination of the negative momentum trick and a novel double sampling scheme, which we called \textit{Sampled Negative Momentum}. We further prove that this accelerated variant achieves the best known oracle complexity for strongly convex problems, which is $\mathcal{O}((n\!+\!\sqrt{\kappa n})\log(1/\epsilon))$.
	\item We discuss some subtle differences on strongly convex assumptions when applying the acceleration technique. Such differences are always neglected in previous directly accelerated methods (e.g., Katyusha and MiG). Our discussion shows that the strongly convex assumption imposed in this paper can be adapted to other strongly convex assumption using a transforming trick.
	\item We provide a variant of the proposed algorithm
	for the non-smooth setting and prove that it achieves a lower $\mathcal{O} \left(\frac{\log(1/\epsilon)}{\sqrt{\epsilon}}\right)$ oracle complexity than the $\mathcal{O}(\frac{1}{\epsilon})$ derived in Point-SAGA~\citep{defazio:sagab}.
	\item Since SSNM does not use the hybrid momentum in Katyusha, it has a simpler structure and potentially clearer intuition. We provide some insights by building connections between the negative momentum trick and the standard Nesterov's momentum in~\citep{nesterov:co}.
\end{itemize}

\section{Preliminaries}

In this paper, we consider Problem~(\ref{prob_def}) in standard Euclidean space with the Euclidean norm denoted by $\norm{\cdot}$. We use $\mathbb{E}$ to denote that the expectation is taken with respect to all randomness in one epoch. In order to further categorize the objective functions, we define that a convex function $f:\R^n \rightarrow \R$ is said to be $L$-smooth if for all $x, y \in \R^d$, it holds that
\begin{equation} \label{l-smooth}
f(x) \leq f(y) + \langle \pf{y}, x-y \rangle + \frac{L}{2} \norm{x-y}^2,
\end{equation}
and $\mu$-strongly convex if for all $x, y \in \R^d$,
\begin{equation} \label{s-convex}
f(x) \geq f(y) + \langle \mathcal{G}, x-y \rangle + \frac{\mu}{2} \norm{x-y}^2,
\end{equation}
where $\mathcal{G}\!\in\!\partial f(y)$, the set of sub-gradient of $f(\cdot)$ at $y$ for non-differentiable $f(\cdot)$. If $f(\cdot)$ is differentiable, we can simply replace $\mathcal{G}\!\in\!\partial f(y)$ with $\mathcal{G}\!=\!\pf{y}$. Then we make the following assumption to identify the main objective condition (strongly convex) that is the focus of this paper:

\begin{sconvex}[Strongly Convex]\label{assump1}
	In Problem~\eqref{prob_def}, each $f_i(\cdot)$\footnote{In fact, if each $f_i(\cdot)$ is $L$-smooth, the averaged function $f(\cdot)$ is itself $L$-smooth --- but probably with a smaller $L$. We keep using $L$ as the smoothness constant for a consistent analysis.} is $L$-smooth and convex, $h(\cdot)$ is $\mu$-strongly convex.
\end{sconvex}

\section{Direct Acceleration of SAGA}
\begin{algorithm*}[t]
	\caption{SAGA with Sampled Negative Momentum (SSNM)}
	\label{SIAM_SC}
	\renewcommand{\algorithmicrequire}{\textbf{Input:}}
	\renewcommand{\algorithmicensure}{\textbf{Initialize:}}
	\begin{algorithmic}[1]
		\REQUIRE Iterations number $K$, initial point $x_1$, learning rate $\eta = \begin{cases}
		\sqrt{\frac{1}{3\mu n L}} &\text{if } \frac{n}{\kappa} \leq \frac{3}{4}, \\
		\frac{1}{2\mu n} & \text{if } \frac{n}{\kappa} > \frac{3}{4}.
		\end{cases}$, parameter $\tau = \frac{n\eta \mu}{1 + \eta\mu}$.\\
		\ENSURE ``Points'' table $\phi$ with $\phi^{1}_{1}=\phi^{1}_{2}=\ldots=\phi^{1}_{n}=x_{1}$ and a running average for the gradients of ``points'' table.\\
		\FOR{$k=1,2,\ldots, K$}
		\STATE {1. Sample $i_k$ uniformly in $\lbrace 1,\ldots,n \rbrace$ and compute the gradient estimator using the running average.}
		\STATE {\ \ \ \ $y^{k}_{i_k}=\tau x_{k} + (1-\tau)\phi_{i_k}^{k}$;} 
		\STATE {\ \ \ \ $\widetilde{\nabla}_k=\nabla f_{i_{k}}(y^{k}_{i_k})-\nabla f_{i_{k}}(\phi^{k}_{i_{k}}) + \frac{1}{n} \sum_{i=1}^n {\pfi{\phi^k_i}}$;}
		\STATE {2. Perform a proximal step.}
		\STATE {\ \ \ \ $x_{k+1}=\arg\min_{x}\left\{h(x)+\langle \widetilde{\nabla}_k, x\rangle+\frac{1}{2\eta}\|x_{k}-x\|^{2}\right\};$}
		\STATE {3. Sample $I_k$ uniformly in $\lbrace 1,\ldots,n \rbrace$ , take $\phi^{k+1}_{I_{k}} =  \tau x_{k+1} + (1 - \tau) \phi^k_{I_k}$. All other entries in the ``points'' table remain unchanged. Update the running average corresponding to the change in the ``points'' table.}
		\ENDFOR
		\renewcommand{\algorithmicensure}{\textbf{Output:}}
		\ENSURE { $x_{K+1}$}
	\end{algorithmic}
\end{algorithm*}

Our proposed algorithm SSNM (SAGA with Sampled Negative Momentum) is formally given in Algorithm~\ref{SIAM_SC}. As we can see, there are some unusual tricks used in Algorithm~\ref{SIAM_SC}. Thus we elaborate some ideas behind Algorithm~\ref{SIAM_SC} by making the following remarks:

\begin{itemize}
	\item \textit{Coupled point $y^k_{i_k}$ correlates to the randomness of $i_k$.} Unlike the negative momentum used for Katyusha, which comes from a fixed snapshot $\tilde{x}$, the negative momentum of SAGA can only be found on a ``points'' table that changes over time. Thus, in SSNM, we choose to use the $i_k$th entry of the ``points'' table to provide the negative momentum, which makes the coupled point correlate to the randomness of sample $i_k$. In fact, all the possible coupled points $y^k_i$ form a ``coupled table''. Although the table is never explicitly computed, we shall see that the concept of ``coupled table'' is critical in the proof of SSNM. The $3$rd step in Algorithm~\ref{SIAM_SC} can thus be regarded as sampling a point in such a table.
	\item \textit{``Biased'' gradient estimator $\tilde{\nabla}_k$.}  The expectation of the semi-stochastic gradient estimator $\tilde{\nabla}_k$ defined in Algorithm~\ref{SIAM_SC} is the average of the gradients computed in the ``coupled table'', $\Eik{\tilde{\nabla}_k} = \frac{1}{n}\sum_{i=1}^n{\pfi{y_i^k}}$, which seems to be surprising as this expectation (except $\tilde{\nabla}_1$) does not correspond to any gradient of $f(\cdot)$, but can be used to show convergence to the optimal solution of $F(\cdot)$. In some sense, $\tilde{\nabla}_k$ is a ``biased'' gradient estimator. 
	\item \textit{Independent samples $I_k$ and $i_k$.} The additional sample $I_k$ is crucial for the convergence analysis of Algorithm~\ref{SIAM_SC}, which chooses an index to store the updated point in the ``points'' table. The insight of this choice is that it separates the randomness of $x_{k+1}$ and the update index in the ``points'' table so as to make certain inequalities valid.
	\item \textit{Two learning rates for two cases.} Using different parameter settings for different objective conditions (ill-condition and well-condition) is common for accelerated methods~\citep{shalev2014accelerated, zhu:Katyusha, kw:MiG}. If some parameters such as $L$, $\mu$ are unknown, SSNM is still a practical algorithm with tuning only $\eta$ and $\tau$, as compared with Katyusha which has 4 parameters that need to be
	tuned. Note that we have tried to make the parameter settings in SSNM similar to Katyusha and MiG. We believe that it can help conduct some fair experimental comparisons with these methods. 
	\item \textit{Only one variable vector with a simple algorithm structure.} Same as MiG in~\citep{kw:MiG}, SSNM only has one variable vector in the main loop. Coupled point $y^k_{i_k}$ can be computed whenever used and does not need to be explicitly stored. Moreover, SSNM has a one loop structure compared to those variants of SVRG. Such a structure is good for asynchronous implementation since algorithms with two loops in this setting always require a synchronization after each inner loop~\citep{man:perturbed}. Moreover, the algorithm structure of SSNM is more elegant than Katyusha and MiG, both of which require a tricky weighted averaged scheme at the end of each inner loop\footnote{These two algorithms can adopt an uniformly average scheme, but in this case, both algorithms require certain restarting tricks, which make them less implementable.}.
\end{itemize}

Since Point-SAGA and SAGA are closely related to SSNM, we compare them in details in Table~\ref{table:comp_SAGA}. SSNM yields the same fast $\mathcal{O}((n+\sqrt{\kappa n}) \log(1/\epsilon))$ convergence rate as Point-SAGA without requiring additional assumptions, demonstrating the advantage of direct acceleration. Note that even for logistic regression, the proximal operator oracle required by Point-SAGA does not have a closed form solution. We may need to run several Newton steps for an inexact oracle as in~\citep{defazio:sagab}. In comparison, the gradient oracle required by SSNM and SAGA is much easier to access. For the memory complexity, as we will discuss in the next subsection, if the objective is some linear models (e.g., loss function with linear predictors), all three methods enjoy an efficient $O(n)$ memory overhead. These aspects demonstrate that SSNM is clearly superior to both SAGA and Point-SAGA.

\begin{table*}[t]
	\centering
	\caption{Comparison of variants of SAGA (All complexities are for strongly convex objectives).}
	\vspace{10pt}
	\begin{tabular}{|l|c|c|c|}
		\hline
		\rule{0pt}{11pt}&Complexity  & Requirements &Memory \\
		\hline
		\rule{0pt}{11pt} SAGA  & $\mathcal{O}((n+\kappa) \log(1/\epsilon))$     & IFO of $f(\cdot)$, PO of $h(\cdot)$ & $O(nd)$ or $O(n)$ for linear models.\\
		\hline
		\rule{0pt}{11pt} Point-SAGA & $\mathcal{O}((n+\sqrt{\kappa n}) \log(1/\epsilon))$   & PO of each $F_i(\cdot)$ &$O(nd)$ or $O(n)$ for linear models$^*$. \\
		\hline
		\rule{0pt}{11pt} SSNM &$\mathcal{O}((n+\sqrt{\kappa n}) \log(1/\epsilon))$  & IFO of $f(\cdot)$, PO of $h(\cdot)$&$O(nd)$ {\color{red} or $O(n)$ for linear models}. \\
		\hline
	\end{tabular}
	
	{\raggedleft\footnotesize{$^*$ A memory issue of Point-SAGA is discussed in Appendix~\ref{sec:memo_point}.} \ \ \ \ \ \ \par}
	\label{table:comp_SAGA}
\end{table*}

\subsection{Implementation}

We discuss the following implementation issues about SSNM:
\begin{itemize}
	\item \textbf{Memory.} For many problems associated with loss minimization of linear predictors (i.e., logistic regression and least squares), we can write each $f_i(x)$ in Problem~\eqref{prob_def} as $\psi_i(\innr{a_i, x})$, where $a_1, \ldots, a_n$ are data vectors. In this case, $\nabla f_i(\phi_i) = \nabla \psi_i(\innr{a_i, \phi_i}) \cdot a_i$ and thus we can reduce the memory consumption of SAGA by storing the scalar $\nabla \psi_i(\innr{a_i, \phi_i})$ instead of the gradient vector. For Point-SAGA, similar trick can be used for objectives with square loss or hinge loss~\citep{defazio:sagab}. However, when an $\ell2$-regularizer is included in each $F_i(\cdot)$, as we point out in Appendix~\ref{sec:memo_point}, the memory overhead of Point-SAGA will always be $O(nd)$. For SSNM, we can reduce the memory complexity by storing the inner product $\innr{a_i, \phi_i}$, and thus SSNM enjoys the same $O(n)$ memory consumption as that of SAGA. We provide the key steps of Algorithm~\ref{SIAM_SC} using this trick here.
	\[
	\begin{aligned}
		&\text{\textbf{Stored:} ``Inner products'' table $\Phi^k$ with $\Phi^k_i = \innr{a_i, \phi^k_i}$ and a running average $\Psi^k$.}\\
		&\textbf{At iteration $k$:} \\
		&\ \ \text{1. Sample $i_k$ uniformly in $\lbrace 1,\ldots,n \rbrace$ and compute the gradient estimator.}\\
		&\ \ \ \ \ \ \innr{a_{i_k}, y^{k}_{i_k}}=\tau \innr{a_{i_k}, x_{k}} + (1-\tau)\Phi^k_{i_k}; \\
		&\ \ \ \ \ \ \widetilde{\nabla}_k= \left(\nabla \psi_{i_k}(\innr{a_{i_k}, y^k_{i_k}})-\nabla \psi_{i_k}(\Phi^k_{i_k}) \right) \cdot a_{i_k} + \Psi^k;\\
		&\ \ \text{2. Perform a proximal update for $x_{k+1}$.}\\
		&\ \ \text{3. Sample $I_k$ uniformly in $\lbrace 1,\ldots,n \rbrace$ , take $\Phi^{k+1}_{I_{k}} =  \tau \innr{a_{I_k}, x_{k+1}} + (1 - \tau) \Phi^k_{I_k}$.}\\
		&\ \ \text{4. Update the running average.} \\
		&\ \ \ \ \ \ \Psi^{k+1} = \Psi^k + \frac{1}{n} \left(\nabla \psi_{I_k}(\Phi^{k+1}_{I_k}) - \nabla \psi_{I_k}(\Phi^k_{I_k}) \right) \cdot a_{I_k};
	\end{aligned}
	\]
	\item \textbf{Per-iteration complexity.} In general, each iteration of SSNM requires computing $4$ stochastic gradients, i.e., $2$ for calculating the gradient estimator and $2$ for updating the running average. In the above case where we use linear predictors, we may consider storing additional $n$ scalars $\nabla \psi_i(\Phi^k_{i})$ to reduce the per-iteration IFO calls to $2$. In comparison, SAGA only computes $1$ stochastic gradient in an iteration.
	\item \textbf{Sparse data vector.} We can use the ``just in time'' update~\citep{roux:sag} or ``lazy/delayed update''~\citep{koneeny:mini} technique for SSNM. The only difference is that in each iteration, we need to consider the coordinates that belong to $support(a_{i_k}) \cup support(a_{I_k})$. We may also use the sparse proximal technique in~\citep{ped:breaking}, which results in a cleaner implementation, but at the expense of potentially losing the accelerated rate as is the case for MiG in~\citep{kw:MiG}.
\end{itemize}

\section{Theory}

In this section, we theoretically analyze the performance of SSNM. First, we give a variance bound of the stochastic gradient estimator of SSNM shown in Lemma~\ref{variance_bound}. Since the stochastic gradient estimator of SSNM is computed at a coupled point that contains randomness, the variance bound for SSNM, unlike most of the variance bounds in previous work, is built with respect to the expectation of the ``biased'' gradient estimator\footnote{Other methods using biased gradient estimators include SARAH~\citep{SARAH}, JacSketch~\citep{JacSketch}}.

\begin{variance_bound} [Variance Bound]
	Using the same notations as in Algorithm~\ref{SIAM_SC}, we can bound the variance of stochastic gradient estimator $\widetilde{\nabla}_k$ as	
	\label{variance_bound}
	\[
		\EikBig{\normbig{\widetilde{\nabla}_k - \frac{1}{n}\sum_{i=1}^n{\pfi{y_i^k}}}^2} \leq 2L\left (\frac{1}{n} \sum_{i=1}^n {\big (f_i(\phi^k_i) - f(y^k_i)\big )} - \frac{1}{n}\sum_{i=1}^n{\innrbig{\pfi{y^k_i}, \phi_{i}^k - y^k_i}}\right ).
	\]
	\begin{proof}
		\[
		\begin{aligned}
			\EikBig{\normbig{\widetilde{\nabla}_k - \frac{1}{n}\sum_{i=1}^n{\pfi{y_i^k}}}^2} &= \EikBig{\Big\lVert\Big (\nabla f_{i_{k}}(y^{k}_{i_k})- \nabla f_{i_{k}}(\phi^{k}_{i_{k}})\Big ) - \frac{1}{n}\sum_{i=1}^n{\big ( \pfi{y_i^k} - \pfi{\phi^k_i}\big )}\Big )\Big\rVert^2} \\
			&\mleq{a} \EikBig{\normbig{\nabla f_{i_{k}}(y^{k}_{i_k})- \nabla f_{i_{k}}(\phi^{k}_{i_{k}})}^2} \\
			&\mleq{b} 2L\cdot \EikBig{f_{i_k}(\phi^k_{i_k}) - f_{i_k}(y^k_{i_k}) - \innrbig{\pfik{y^k_{i_k}}, \phi^k_{i_k} - y^k_{i_k}}} \\
			&= 2L\left (\frac{1}{n} \sum_{i=1}^n {\big (f_i(\phi^k_i) - f(y^k_i)\big )} - \frac{1}{n}\sum_{i=1}^n{\innrbig{\pfi{y^k_i}, \phi_{i}^k - y^k_i}}\right ),
		\end{aligned}
		\]
		where $\mar{a}$ follows from $\E{\norm{\zeta - \mathbb{E}\zeta}^2} \leq \mathbb{E}\norm{\zeta}^2$ and $\mar{b}$ uses Theorem 2.1.5 in~\citep{nesterov:co}.
	\end{proof}
\end{variance_bound}

Now we can formally present the main theorem of SSNM below. As stated in~\citep{zhu:Katyusha}, the major task of the negative momentum is to cancel the additional inner product term shown in the variance bound so as to keep a close connection in each iteration. As we shall see shortly, our proposed sampled negative momentum effectively cancels the inner product term, which is where the acceleration comes from. 

\begin{siam_nsc}
	\label{thm:strongly_convex}
	Let $\xs$ be the solution of Problem~\eqref{prob_def}, define the following Lyapunov function $T$, which is the same as the one in SAGA~\citep{defazio:saga}:
	\[
		T^k \triangleq T(x_k, \phi^k)\triangleq \frac{1}{n\eta\mu} \left(\frac{1}{n}\sum_{i = 1}^n{F_i(\phi^k_i)} - F(\xs) - \frac{1}{n}\sum_{i= 1}^{n} \innr{\pFi{\xs}, \phi^k_i - \xs}\right) + \frac{1}{2\eta n} \norm{x_{k}-\xs}^2.
	\]
	If Assumption~\ref{assump1} holds, then by choosing $\tau = \frac{n\eta\mu}{1 + \eta\mu}$, steps of Algorithm~\ref{SIAM_SC} satisfy the following contraction for the Lyapunov function in expectation (conditional on $T^k$):
	\[
		\Er{i_k, I_k}{T^{k+1}} \leq (1 + \eta \mu)^{-1} T^{k}.
	\]
	Thus, by carefully choosing $\eta$, we have the following inequalities in two cases:
	
	\textup{(I) (For ill-conditioned problems).} If $\frac{n}{\kappa} \leq \frac{3}{4}$, with $\eta = \sqrt{\frac{1}{3\mu n L}}$ it holds that
	\[
		\E{\norm{x_{K+1} - \xs}^2} \leq \left(1 + \sqrt{\frac{1}{3n\kappa}}\right)^{-K}\left (\frac{2}{\mu} \left (F(x_1) - F(\xs)\right ) + \norm{x_{1} - \xs}^2\right ).
	\]
	
	The above inequality implies that in order to reduce the squared norm distance to $\epsilon$, we have an $\mathcal{O}(\sqrt{\kappa n}\log(1/\epsilon))$ oracle complexity as $\epsilon \rightarrow 0$ in expectation. 
		
	\textup{(II) (For well-conditioned problems).} If $\frac{n}{\kappa} > \frac{3}{4}$, by choosing $\eta = \frac{1}{2\mu n}$, we have
	\[
		\E{\norm{x_{K+1} - \xs}^2} \leq \left (1 + \frac{1}{2n} \right )^{-K} \left (\frac{2}{\mu} \big (F(x_1) - F(\xs)\big ) + \norm{x_1 - \xs}^2\right ).
	\]

	This inequality implies that in this case we have an $\mathcal{O}(n\log(1/\epsilon))$ oracle complexity as $\epsilon \rightarrow 0$ in expectation. 
\end{siam_nsc}

Thus, for strongly convex objectives, SSNM yields a fast $\mathcal{O}((n + \sqrt{\kappa n}) \log(1/\epsilon))$, which keeps up with the best known oracle complexity achieved by accelerated SVRG~\citep{roy:appa,zhu:Katyusha}.

\subsection{Proof of Theorem~\ref{thm:strongly_convex}}

The proof combines the ideas in SAGA~\citep{defazio:saga}, Katyusha~\citep{zhu:Katyusha} and~\citep{kw:MiG}.

In order to prove Theorem~\ref{thm:strongly_convex}, we need the following useful lemma, which can be regarded as using the 3-point equality of Bregman divergence in the Euclidean norm setting:

\begin{sc_prox}\label{h_prox}
	If two vectors $x_{k+1}$, $x_{k}\in \R^d$ satisfy $x_{k+1} = \arg\min_{x}\{h(x)+\langle \widetilde{\nabla}_k, x\rangle+\frac{1}{2\eta}\norm{x_{k}-x}^{2}\}$ with a constant vector $\widetilde{\nabla}_k$ and a $\mu$-strongly convex function $h(\cdot)$, then for all $u\in \R^d$, we have
	\[
	\langle \widetilde{\nabla}_k, x_{k+1} - u \rangle \leq -\frac{1}{2\eta}\norm{x_{k+1} - x_{k}}^2 + \frac{1}{2\eta}\norm{x_{k} - u}^2 - \frac{1+\eta\mu}{2\eta}\norm{x_{k+1}-u}^2 + h(u) - h(x_{k+1}).
	\]
\end{sc_prox}
This Lemma is identical to Lemma 3.5 in~\citep{zhu:Katyusha}, and hence the proof is omitted.

First, we analyze Algorithm~\ref{SIAM_SC} at the $k$th iteration, given that the randomness from previous iterations are fixed. 

We start with the convexity of $f_{i_k}(\cdot)$ at $(y^{k}_{i_k}, \xs)$. By definition, we have
\[
\begin{aligned}
f_{i_k} (y^k_{i_k}) - f_{i_k} (\xs) &\leq \innr{\pfik{y^k_{i_k}}, y^k_{i_k} - \xs} \\
&\meq{\star} \frac{1 - \tau}{\tau} \innr{\pfik{y^k_{i_k}}, \phi^k_{i_k} - y^k_{i_k}} + \innr{\pfik{y^k_{i_k}} - \widetilde{\nabla}_k, x_k - \xs}  + \innr{\widetilde{\nabla}_k, x_k - x_{k+1}} \\
&\ \ \ \ + \innr{\widetilde{\nabla}_k, x_{k+1} - \xs},
\end{aligned}
\]
where $\mar{\star}$ uses the definition of the $i_k$th entry of ``coupled table'' that $y^k_{i_k} = \tau x_k + (1 - \tau) \phi^k_{i_k}$.

As we will see, the first term on the right side is used to cancel the unwanted inner product term in the variance bound.

By taking expectation with respect to sample $i_k$ and using the unbiasedness that $\Eik{\pfik{y^k_{i_k}} - \widetilde{\nabla}_k} = \mathbf{0}
$, we obtain
\begin{equation}\label{P1_T1}
\frac{1}{n} \sum_{i=1}^n{f_{i} (y^k_{i})} - f(\xs)
\leq \frac{1 - \tau}{\tau n} \sum_{i=1}^n{\innr{\pfi{y^k_{i}}, \phi^k_{i} - y^k_{i}}} + \Eik{\innr{\widetilde{\nabla}_k, x_k - x_{k+1}}} + \Eik{\innr{\widetilde{\nabla}_k, x_{k+1} - \xs}}.
\end{equation}

In order to bound $\Eik{\innr{\widetilde{\nabla}_k, x_k - x_{k+1}}}$, we use the $L$-smoothness of $f_{I_k}(\cdot)$ at $(\phi^{k+1}_{I_k}, y^k_{I_k})$ , which is
\[
\begin{aligned}
f_{I_k}(\phi^{k+1}_{I_k}) - f_{I_k} (y^k_{I_k}) &\leq \innr{\pfr{I_k}{y^k_{I_k}}, \phi^{k+1}_{I_k} - y^k_{I_k}} + \frac{L}{2} \norm{\phi^{k+1}_{I_k} - y^k_{I_k}}^2.
\end{aligned}
\]

Taking expectation with respect to sample $I_k$ and using our choice of $\phi^{k+1}_{I_k} = \tau x_{k+1} + (1 - \tau)\phi^k_{I_k}$ as well as the definition of ``coupled table'', we conclude that
\[
\begin{gathered}
\Er{I_k}{f_{I_k}(\phi^{k+1}_{I_k})} - \frac{1}{n}\sum_{i = 1}^n{f_{i} (y^k_{i})} \leq \tau\innrbig{\frac{1}{n}\sum_{i=1}^n{\pfi{y^k_{i}}}, x_{k+1} - x_k} + \frac{L\tau^2}{2} \norm{x_{k+1} - x_k}^2, \\
\innr{\widetilde{\nabla}_k, x_k - x_{k+1}} \leq \frac{1}{\tau n}\sum_{i = 1}^n {f_{i} (y^k_{i})} -  \frac{1}{\tau}\Er{I_k}{f_{I_k}(\phi^{k+1}_{I_k})} + \innrbig{\frac{1}{n}\sum_{i=1}^n{\pfi{y^k_{i}}} - \widetilde{\nabla}_k, x_{k+1} - x_k} + \frac{L\tau}{2} \norm{x_{k+1} - x_k}^2.
\end{gathered}
\]

Here we see the effect of the independent sample $I_k$. It decouples the randomness of $x_{k+1}$ and the update position so as to make the above inequalities valid.

Taking expectation with respect to sample $i_k$, we obtain
\begin{align}
\Eik{\innr{\widetilde{\nabla}_k, x_k - x_{k+1}}} &\leq \frac{1}{\tau n}\sum_{i = 1}^n {f_{i} (y^k_{i})} -  \frac{1}{\tau}\Er{i_k, I_k}{f_{I_k}(\phi^{k+1}_{I_k})} + \EikBig{\innrbig{\frac{1}{n}\sum_{i=1}^n{\pfi{y^k_{i}}} - \widetilde{\nabla}_k, x_{k+1} - x_k}} \nonumber\\
&\ \ \ \ + \frac{L\tau}{2} \Eik{\norm{x_{k+1} - x_k}^2}. \label{P1_T2}
\end{align}

By upper bounding~\eqref{P1_T1} using~\eqref{P1_T2} and Lemma~\ref{h_prox} (with $h(\cdot)$ $\mu$-strongly convex and $u = \xs$), we obtain

\[
\begin{aligned}
\frac{1}{n} \sum_{i=1}^n{f_{i} (y^k_{i})} - f(\xs)
&\leq \frac{1 - \tau}{\tau n} \sum_{i=1}^n{\innr{\pfi{y^k_{i}}, \phi^k_{i} - y^k_{i}}} + \frac{1}{\tau n}\sum_{i = 1}^n {f_{i} (y^k_{i})} -  \frac{1}{\tau}\Er{i_k, I_k}{f_{I_k}(\phi^{k+1}_{I_k})} \\
&\ \ \ \ + \EikBig{\innrbig{\frac{1}{n}\sum_{i=1}^n{\pfi{y^k_{i}}} - \widetilde{\nabla}_k, x_{k+1} - x_k}} + \frac{L\tau}{2} \Eik{\norm{x_{k+1} - x_k}^2} \\
&\ \ \ \ -\frac{1}{2\eta}\Eik{\norm{x_{k+1} - x_{k}}^2} + \frac{1}{2\eta}\norm{x_{k} - \xs}^2 - \frac{1+\eta\mu}{2\eta}\Eik{\norm{x_{k+1}-\xs}^2}\\
&\ \ \ \  + h(\xs) - \Eik{h(x_{k+1})}.
\end{aligned}
\]

Here we add a constraint that $L\tau \leq \frac{1}{\eta} - \frac{L\tau}{1 - \tau}$, which is identical to the one used in~\citep{kw:MiG}. Using Young's inequality $\innr{a, b} \leq \frac{1}{2\beta} \norm{a}^2 + \frac{\beta}{2}\norm{b}^2$ to upper bound $\EikBig{\innr{\frac{1}{n}\sum_{i=1}^n{\pfi{y^k_{i}}} - \widetilde{\nabla}_k, x_{k+1} - x_k}}$ with $\beta = \frac{L\tau}{1 - \tau}>0$, we can simplify the above inequality as
\[
\begin{aligned}
\frac{1}{n} \sum_{i=1}^n{f_{i} (y^k_{i})} - f(\xs)
&\leq \frac{1 - \tau}{\tau n} \sum_{i=1}^n{\innr{\pfi{y^k_{i}}, \phi^k_{i} - y^k_{i}}} + \frac{1}{\tau n}\sum_{i = 1}^n {f_{i} (y^k_{i})} -  \frac{1}{\tau}\Er{i_k, I_k}{f_{I_k}(\phi^{k+1}_{I_k})} \\
&\ \ \ \ + \frac{1 - \tau}{2L\tau}\EikBig{\normbig{\frac{1}{n}\sum_{i=1}^n{\pfi{y^k_{i}}} - \widetilde{\nabla}_k}^2} + \frac{1}{2\eta}\norm{x_{k} - \xs}^2 - \frac{1+\eta\mu}{2\eta}\Eik{\norm{x_{k+1}-\xs}^2}\\
&\ \ \ \  + h(\xs) - \Eik{h(x_{k+1})}.
\end{aligned}
\]

By applying Lemma~\ref{variance_bound} to upper bound the variance term, we see that the additional variance term in the variance bound is canceled by the sampled momentum, which gives
\begin{align}
\frac{1}{n} \sum_{i=1}^n{f_{i} (y^k_{i})} - f(\xs)
&\leq \frac{1}{\tau n}\sum_{i = 1}^n {f_{i} (y^k_{i})} -  \frac{1}{\tau}\Er{i_k, I_k}{f_{I_k}(\phi^{k+1}_{I_k})} + \frac{1 - \tau}{\tau n} \sum_{i=1}^n {\big (f_i(\phi^k_i) - f(y^k_i)\big )}\nonumber\\
&\ \ \ \  + \frac{1}{2\eta}\norm{x_{k} - \xs}^2 - \frac{1+\eta\mu}{2\eta}\Eik{\norm{x_{k+1}-\xs}^2} + h(\xs) - \Eik{h(x_{k+1})},\nonumber\\
\frac{1}{\tau}\Er{i_k, I_k}{f_{I_k}(\phi^{k+1}_{I_k})} - F(\xs)
&\leq \frac{1 - \tau}{\tau n} \sum_{i=1}^n {f_i(\phi^k_i)}  + \frac{1}{2\eta}\norm{x_{k} - \xs}^2 - \frac{1+\eta\mu}{2\eta}\Eik{\norm{x_{k+1}-\xs}^2} - \Eik{h(x_{k+1})}.\label{P1_T3}
\end{align}

Using the convexity of $h(\cdot)$ and that $\phi_{I_k}^{k+1} = \tau x_{k+1} + (1 - \tau) \phi_{I_k}^k$, we have
\[
\begin{aligned}
h(\phi^{k+1}_{I_k}) &\leq \tau h(x_{k+1}) + (1 - \tau) h(\phi^k_{I_k}).
\end{aligned}
\]

After taking expectation with respect to sample $I_k$ and sample $i_k$, we obtain
\[
-\Eik{h(x_{k+1})} \leq \frac{1 - \tau}{\tau n} \sum_{i = 1}^n{h(\phi^k_{i})} - \frac{1}{\tau} \Er{i_k, I_k}{h(\phi^{k+1}_{I_k})}.
\]

Combining the above inequality with~\eqref{P1_T3} and using  the definition that $F_i(\cdot) = f_i(\cdot) + h(\cdot)$, we can write~\eqref{P1_T3} as
\[
\frac{1}{\tau} \Er{i_k, I_k}{F_{I_k} (\phi_{I_k}^{k+1}) - F_{I_k}(\xs)} \leq \frac{1-\tau}{\tau}\Big (\frac{1}{n}\sum_{i = 1}^n{F_i(\phi^k_i)} - F(\xs)\Big ) + \frac{1}{2\eta}\norm{x_{k} - \xs}^2 - \frac{1+\eta\mu}{2\eta}\Eik{\norm{x_{k+1}-\xs}^2}.
\]

Dividing the above inequality by $n$ and adding both sides by $\frac{1}{\tau n}\Er{I_k}{\sum_{i\neq I_k}^n {\big (F_i(\phi^{k}_i) - F_i(\xs)\big )}}$, we obtain
\begin{align}
\frac{1}{\tau} \ErBig{i_k, I_k}{\frac{1}{n}\sum_{i=1}^n{F_{i} (\phi_{i}^{k+1})} - F(\xs)} &\leq \frac{1-\tau}{\tau n} \Big (\frac{1}{n}\sum_{i = 1}^n{\big (F_i(\phi^k_i) - F_i(\xs)\big )}\Big ) + \frac{1}{\tau n}\ErBig{I_k}{\sum_{i\neq I_k}^n {\big (F_i(\phi^{k}_i) - F_i(\xs)\big )}} \nonumber\\
&\ \ \ \ + \frac{1}{2\eta n}\norm{x_{k} - \xs}^2 - \frac{1+\eta\mu}{2\eta n}\Eik{\norm{x_{k+1}-\xs}^2} \nonumber\\
&= \frac{1-\tau}{\tau n} \Big (\frac{1}{n}\sum_{i = 1}^n{\big (F_i(\phi^k_i) - F_i(\xs)\big )}\Big ) + \frac{1}{\tau n^2}\sum_{j=1}^n{\sum_{i\neq j}^n {\big (F_i(\phi^{k}_i) - F_i(\xs)\big )}} \nonumber\\
&\ \ \ \ + \frac{1}{2\eta n}\norm{x_{k} - \xs}^2 - \frac{1+\eta\mu}{2\eta n}\Eik{\norm{x_{k+1}-\xs}^2} \nonumber\\
&= \frac{1 - \frac{\tau}{n}}{\tau} \Big (\frac{1}{n}\sum_{i = 1}^n{F_i(\phi^k_i)} - F(\xs)\Big ) + \frac{1}{2\eta n}\norm{x_{k} - \xs}^2 \label{P1_T6} \\
&\ \ \ \  - \frac{1+\eta\mu}{2\eta n}\Eik{\norm{x_{k+1}-\xs}^2}.\nonumber
\end{align}

Since $\frac{1}{n}\sum_{i = 1}^n{F_i(\phi^k_i) - F(\xs)}$ may not be positive, we need to involve the following term in our Lyapunov function:
\[
\begin{aligned}
-\frac{1}{n} \sum_{i=1}^n{\innr{\pFi{\xs}, \phi^{k+1}_i - \xs}} &= -\frac{1}{n} \innr{\pFr{I_k}{\xs}, \phi^{k+1}_{I_k} - \xs} - \frac{1}{n} \sum_{i\neq I_k}^{n} \innr{\pFi{\xs}, \phi^k_i - \xs} \\	
&= -\frac{\tau}{n} \innr{\pFr{I_k}{\xs},  x_{k+1}- \xs} + \frac{\tau}{n} \innr{\pFr{I_k}{\xs},  \phi^k_{I_k}- \xs} \\
&\ \ \ \ - \frac{1}{n} \sum_{i= 1}^{n} \innr{\pFi{\xs}, \phi^k_i - \xs}.
\end{aligned}
\]

After taking expectation with respect to sample $I_k$ and $i_k$, we obtain
\begin{equation}\label{P1_T5}
\ErBig{i_k, I_k}{-\frac{1}{n} \sum_{i=1}^n{\innr{\pFi{\xs}, \phi^{k+1}_i - \xs}}} = - \left(1 - \frac{\tau}{n}\right) \Big(\frac{1}{n}\sum_{i= 1}^{n} \innr{\pFi{\xs}, \phi^k_i - \xs}\Big).
\end{equation}

In order to give a clean proof, we denote $D_k \triangleq \frac{1}{n}\sum_{i = 1}^n{F_i(\phi^k_i)} - F(\xs)-\frac{1}{n}\sum_{i= 1}^{n} \innr{\pFi{\xs}, \phi^k_i - \xs}$ and $P_k \triangleq \norm{x_{k}-\xs}^2$, then by combining~\eqref{P1_T6},~\eqref{P1_T5}, we can write the contraction as
\begin{equation} \label{P1_T4}
\frac{1}{\tau} \Er{i_k, I_k}{D_{k+1}} + \frac{1 + \eta \mu}{2\eta n} \Eik{P_{k+1}} \leq \frac{1 - \frac{\tau}{n}}{\tau} D_k + \frac{1}{2\eta n} P_k.
\end{equation}

\textbf{Case I:} Consider the first case with $\frac{n}{\kappa} \leq \frac{3}{4}$, choosing $\eta = \sqrt{\frac{1}{3\mu n L}}$ and $\tau = \frac{n\eta \mu}{1 + \eta\mu} = \frac{\sqrt{\frac{n}{3\kappa}}}{1 + \sqrt{\frac{1}{3n\kappa}}} < \frac{1}{2}$, we first evaluate the parameter constraint:
\[
L\tau \leq \frac{1}{\eta} - \frac{L\tau}{1 - \tau} \Rightarrow \underbrace{\frac{2 - \tau}{1 - \tau}}_{< 3} \cdot \underbrace{\frac{\sqrt{\frac{n}{3\kappa}}}{1 + \sqrt{\frac{1}{3n\kappa}}}}_{\leq \sqrt{\frac{n}{3\kappa}}} \leq \sqrt{\frac{3n}{\kappa}},
\]
which means that the constraint is satisfied by our parameter choices.

Moreover, with this choice of $\tau$, we have
\[
\frac{1}{\tau (1 + \eta \mu)} = \frac{1 - \frac{\tau}{n}}{\tau} = \frac{1}{n\eta\mu}.
\]

Thus, the contraction~\eqref{P1_T4} can be written as
\[
\frac{1}{n\eta\mu} \Er{i_k, I_k}{D_{k+1}} + \frac{1}{2\eta n} \Eik{P_{k+1}} \leq (1 + \eta \mu)^{-1}\cdot \Big (\frac{1}{n\eta\mu} D_k + \frac{1}{2\eta n} P_k\Big ).
\]

After telescoping the above contraction from $k = 1\ldots K$ and taking expectation with respect to all randomness, we have
\[
\frac{1}{n\eta\mu} \E{D_{K+1}} + \frac{1}{2\eta n} \E{P_{K+1}} \leq (1 + \eta \mu)^{-K}\cdot \Big (\frac{1}{n\eta\mu} D_1 + \frac{1}{2\eta n} P_1 \Big ).
\]

Note that $D_1 = F(x_1) - F(\xs)$ and $\E{D_{K+1}} \geq 0$ based on convexity. After substituting the parameter choices, we have
\[
\E{\norm{x_{K+1} - \xs}^2} \leq \Big (1 + \sqrt{\frac{1}{3n\kappa}}\Big )^{-K}\cdot\Big (\frac{2}{\mu} \big (F(x_1) - F(\xs)\big ) + \norm{x_{1} - \xs}^2\Big ).
\]

\textbf{Case II:} Consider another case with $\frac{n}{\kappa} > \frac{3}{4}$, choosing $\eta = \frac{1}{2\mu n}$, $\tau =  \frac{n\eta \mu}{1 + \eta\mu} = \frac{\frac{1}{2}}{1 + \frac{1}{2n}} < \frac{1}{2}$. Again, we first evaluate the constraint:
\[
L\tau \leq \frac{1}{\eta} - \frac{L\tau}{1 - \tau} \Rightarrow \tau \cdot \underbrace{\frac{2 - \tau}{1 - \tau}}_{< 3} < \frac{3}{2} < \frac{2n}{\kappa}.
\]

Then by rewriting the contraction~\eqref{P1_T4}, telescoping from $k = 1\ldots K$ and taking expectation with respect to all randomness, we obtain
\[
2 \E{D_{K+1}} + \frac{1}{2\eta n} \E{P_{K+1}} \leq (1 + \eta\mu)^{-K} \cdot \Big (2 D_1 + \frac{1}{2\eta n} P_1\Big ).
\]

By substituting the parameter choices, we have
\[
\E{\norm{x_{K+1} - \xs}^2} \leq \Big (1 + \frac{1}{2n} \Big )^{-K} \cdot \Big (\frac{2}{\mu} \big (F(x_1) - F(\xs)\big ) + \norm{x_1 - \xs}^2\Big ).
\]

\subsection{Some subtle differences on strongly convex assumption}
\label{sec:subtle}

Recall that the strongly convex assumption for SAGA is imposed on each $f_i(\cdot)$ (or the average $f(\cdot)$ as an extension)~\citep{defazio:saga}. In comparison, SSNM requires the strong convexity of $h(\cdot)$ (in Assumption~\ref{assump1}), which seems to be critical in the proof. Below we show that the strong convexity assumption of each $f_i(\cdot)$ can be efficiently transformed into Assumption~\ref{assump1}.

\noindent\textbf{Transforming the strong convexity assumption from holding for all $f_i(\cdot)$ to Assumption~\ref{assump1}:} Suppose we have an objective in the form~\eqref{prob_def} with each $f_i(\cdot)$ $L$-smooth and $\mu$-strongly convex, $h(\cdot)$ convex and proper (the main assumption of SAGA). By defining $f'_i(\cdot) = f_i(\cdot) - \frac{\mu}{2} \norm{\cdot}^2$ for each $f_i(\cdot)$ and $h'(\cdot) = h(\cdot) + \frac{\mu}{2} \norm{\cdot}^2$, the optimal solution of minimizing $F'(\cdot) = \frac{1}{n} \sum_{i=1}^n{f'_i(\cdot)} + h'(\cdot)$ is equivalent to that of~\eqref{prob_def} and it can be verified that each $f'_i(\cdot)$ is $(L - \mu)$-smooth and convex, $h'(\cdot)$ is $\mu$-strongly convex. Moreover, the proximal operator $\prox^{\eta}_{h'}(v) \triangleq \arg\min_{x}\{h'(x) + \frac{1}{2\eta} \norm{x - v}^2\}, \forall v\in \R^d$ can be efficiently computed as
\[
\text{$\prox$}^{\eta}_{h'}(v) = \text{$\prox$}^{\eta / (1 + \eta\mu)}_{h}\left(\frac{v}{1 + \eta\mu}\right).
\]

Conversely, Assumption~\ref{prob_def} may not be reducible to the strong convexity assumption of each $f_i(\cdot)$ using the above trick, since the modified regularizer $h(\cdot) - \frac{\mu}{2} \norm{\cdot}^2$ may not be as ``proper'' as $h(\cdot)$. 

Directly accelerated variants of SVRG (e.g., Katyusha and MiG) also require a strongly convex regularizer to achieve acceleration. This requirement can be weakened by adopting a restarting scheme for MiG (Algorithm~3 with Option~II in \citep{kw:MiG})~\footnote{Similar restarting trick can be used for Katyusha to weaken the strongly convex assumption.}, which only requires $F(\cdot)$ to be strongly convex and thus keeps the same assumption as in Prox-SVRG~\citep{xiao:prox-svrg}. Unfortunately, we found that the similar trick does not work for SSNM. The best we can achieve is to slightly weaken the strong convexity assumption to be imposed on each $F_i(\cdot)$, but it requires an additional upper bound $F(x) - F(\xs) \leq \frac{L_F}{2} \norm{x - \xs}^2$ for all $x\in \R^d$, where $L_F$ is potentially much larger than $L$ ($L_F = L$ when $h(\cdot) \equiv 0$). Moreover, the algorithm structure will be more complicated than Algorithm~\ref{SIAM_SC}. Thus, we decided not to include the variant here.

\subsection{Non-smooth extension}

Problem~\eqref{prob_def} with non-smooth but $L_1$-Lipschitz continuous $f_i(\cdot)$, strongly convex $h(\cdot)$ is also prevalent in machine learning, e.g., L2-SVM. To solve this type of problems, the most direct solution is using sub-gradient methods (e.g., Pegasos~\citep{pegasos} with an~$\mathcal{O}(\frac{1}{\epsilon})$ rate). As an accelerated variant of SAGA, Point-SAGA also obtains an $\mathcal{O}(\frac{1}{\epsilon})$ rate for a similar type of objectives~\citep{defazio:sagab}. In comparison, Point-SAGA requires the exact proximal operator of each $f_i(\cdot)$ but does not show improvement on the bound. In this subsection, we consider extending SSNM into this setting by utilizing the proximal information of each $f_i(\cdot)$, which results in a convergence rate faster than $\mathcal{O}(\frac{1}{\epsilon})$.

Following~\citep{Prisma}, we apply \textit{Moreau-Yosida regularization} for each $f_i(\cdot)$, which results in a smooth approximation $f^\beta_i(\cdot)$ (with $\beta > 0$) defined as
\[
\forall v \in \R^d, f^{\beta}_i(v) = \inf_{x\in\R^d} \left\{f_i(x) + \frac{1}{2\beta} \norm{x - v}^2\right\}.
\]
Then, it is clear that $\prox^{\beta}_{f_i}(v)$ returns the point that attains the infimum in $f^{\beta}_i(v)$. As proven in Proposition 12.29~\citep{prox-smooth}, $f^{\beta}_i(\cdot)$ is $\frac{1}{\beta}$-smooth and its gradient can be computed as $\nabla f^{\beta}_i(x) = \frac{1}{\beta} (x - \prox^{\beta}_{f_i}(x)), \forall x \in \R^d$. Moreover, we have the following properties to further bound the error in this smooth approximation:
\begin{moreau_envelope}[Lemma 2.2,~\citep{Prisma}]
	\label{eb-smooth}
	Let $f_i(\cdot)$ be an $L_1$-Lipschitz continuous and convex function, then for any $x\in \R^d$, $\beta > 0$
	\[
	f^{\beta}_i(x) \leq f_i(x) \leq f^{\beta}_i(x) + \frac{\beta L_1^2}{2}.
	\]
\end{moreau_envelope}

Thus, by defining a ``smoothed'' objective $F^\beta(\cdot) = \frac{1}{n} \sum_{i=1}^n{f^\beta_i(\cdot)} + h(\cdot)$, we can use SSNM to minimize $F^\beta(\cdot)$, which leads to the following corollary: 
\begin{siam_ns}
	Using Algorithm~\ref{assump1} to minimize $F^\beta(\cdot)$ defined above, and by choosing $\beta = \frac{\mu\epsilon}{4L_1^2}$, where $\epsilon > 0$ (small enough) is the required accuracy, in order to achieve $\norm{x_{K+1} - \xs}^2 \leq \epsilon$ at the output point $x_{K+1}$, where $\xs$ is the solution of minimizing the original $F(\cdot)$, we need an $\mathcal{O}\left(\left(n + \frac{\sqrt{n}L_1}{\sqrt{\epsilon} \mu}\right)\log(1/\epsilon)\right)$ oracle complexity in expectation.
	\begin{proof}
		Denote the optimal solution of minimizing $F^\beta(\cdot)$ as $x^{\star}_{\beta}$. With the strong convexity of $F(\cdot)$, we can bound the difference between $x^{\star}_{\beta}$ and $\xs$ as
		\[
		\begin{aligned}
		\norm{x^{\star}_\beta - \xs}^2 \leq \frac{2}{\mu} \big(F(x^{\star}_{\beta}) - F(\xs)\big).
		\end{aligned}
		\]
		Based on Lemma~\ref{eb-smooth}, we have the following inequalities:
		\[
		F(x^{\star}_\beta) \leq F^{\beta}(x^{\star}_\beta) + \frac{\mu\epsilon}{8} \mleq{\star} F^\beta(\xs) + \frac{\mu\epsilon}{8} \leq F(\xs) + \frac{\mu\epsilon}{8},
		\] 
		where $\mar{\star}$ holds due to the optimality of $x^{\star}_{\beta}$. 
		
		Thus, we conclude that $\norm{x^{\star}_\beta - \xs}^2 \leq \frac{\epsilon}{4}$, which is based on the choice of $\beta$.
		
		Following Theorem~\ref{thm:strongly_convex}, in order to reduce the squared norm distance $\norm{x_{K+1} - x^{\star}_{\beta}}^2$ at the output point $x_{K+1}$ to $\frac{\epsilon}{4}$, we need $\mathcal{O} \left(\left(n + \sqrt{\frac{n}{\beta \mu}}\right)\log(1 / \epsilon)\right)$ oracle calls. Note that the above results imply that $x_{K+1}$ satisfies
		\[
		\begin{aligned}
		\norm{x_{K+1} - \xs}^2 \leq 2\norm{x_{K+1} - x^{\star}_\beta}^2 + 2\norm{x^{\star}_\beta - \xs}^2 \leq \epsilon. 
		\end{aligned}
		\]
	\end{proof}
\end{siam_ns}

The above results imply an $\mathcal{O}\left(\frac{\log(1/\epsilon)}{\sqrt{\epsilon}}\right)$ bound to solve the non-smooth objectives, which is superior to the $\mathcal{O}(\frac{1}{\epsilon})$ obtained by Point-SAGA. In order to avoid the log factor in the bound, we can use the \textsf{AdaptSmooth} in~\citep{zhu:box}. However, as mentioned in Section~\ref{sec:subtle}, in order to satisfy the $\textsf{HOOD}$ property in~\citep{zhu:box}, we need an additional upper bound $F(x) - F(\xs) \leq \frac{L_F}{2} \norm{x - \xs}^2$ for all $x\in \R^d$, which rules out certain choices of $h(\cdot)$, such as the indicator function of a closed convex set. Moreover, a $\log(L_F / \mu)$ factor will appear in the oracle complexity bound after using the \textsf{AdaptSmooth}. Thus, we omit further discussions about eliminating the log factor here.

\section{Some insights about the negative momentum trick}

In~\citep{zhu:Katyusha}, the negative momentum (or Katyusha momentum) is described as a ``magnet'' that reduces the error of the  semi-stochastic gradient estimator for variance reduced algorithms. Thus, the author combined this idea with Nesterov's momentum (or ``positive'' momentum) to achieve acceleration. However, as shown in~\citep{kw:MiG} as well as this work, it seems that merely using the negative momentum trick is enough to obtain the same accelerated convergence rate, which makes this acceleration somewhat ``counter-intuitive''. In theory, it is clear that with the help of negative momentum, we can adopt a much tighter variance bound. However, this theoretical effect does not explain the source of acceleration. In this section, we try to build a connection between the negative momentum and the standard Nesterov's momentum in~\citep{nesterov:co}.

For simplicity, we mainly focus on the objective~\eqref{prob_def} with $h(\cdot)\equiv 0$ in this section. First, consider the deterministic case with $n=1$, Algorithm~\ref{SIAM_SC} degenerates into an algorithm with the following key steps (with $z\in \R^d$ denoting the one item ``points'' table $\phi$):
\[
\begin{gathered}
y_k = \tau x_k + (1 - \tau) z_k; \\
x_{k+1} = x_k - \eta\cdot \pf{y_k}; \\
z_{k+1} = \tau x_{k+1} + (1 - \tau) z_k.
\end{gathered}
\]
This is exactly the scheme of IGA~\citep{interior} in the Euclidean setting. Note that we can completely eliminate the sequence $\{x_k\}$, which results in a simple scheme below.
\[
\begin{gathered}
z_{k+1} = y_k - \eta\tau\cdot \pf{y_k}; \\
y_{k+1} = z_{k+1} + (1 - \tau) (z_{k+1} - z_k).
\end{gathered}
\]
By carefully choosing parameters $\eta$ and $\tau$, we recover the original Nesterov's accelerated gradient method with constant stepsize~\citep{nesterov:co}. This observation motivates us to formulate the key steps in SSNM (Algorithm~\ref{SIAM_SC}) and MiG\footnote{We adopt the uniform averaged scheme of MiG (Algorithm 3 with Option II in~\citep{kw:MiG}) for simplicity.} into the following schemes (outer loops are omitted for simplicity):

\begin{center}
	\begin{minipage}[t]{8cm}
		\begin{center}\textbf{SSNM}\end{center}
		\vspace{-15pt}
		\rule{8cm}{1pt}
		\vspace{-8pt}
		\[
		\begin{aligned}
			&\widetilde{\nabla}^{(1)}_k=\nabla f_{i_{k}}(y^{k}_{i_k})-\nabla f_{i_{k}}(\phi^{k}_{i_{k}}) + \frac{1}{n} \sum_{i=1}^n {\pfi{\phi^k_i}}; \\
			&\phi_{I_k}^{k+1} = y_{I_k}^k - \eta\tau \cdot \widetilde{\nabla}^{(1)}_k;\\[1em]
			&y_{i_{k+1}}^{k+1} = \phi^{k+1}_{I_k} + \underline{(1 - \tau) (\phi^{k+1}_{i_{k+1}} - \phi^k_{I_k})}; \\
		\end{aligned}
		\]
	\end{minipage}
	\begin{minipage}[t]{8cm}
		\begin{center}\textbf{MiG}\end{center}
		\vspace{-15pt}
		\rule{8cm}{1pt}
		\[
		\begin{aligned}
			&\text{for }k=1\ldots m: \\
			&\ \ \ \ \ \ \widetilde{\nabla}^{(2)}_{k} = \pfik{y^s_{k}} - \pfik{\tilde{x}_s} + \pf{\tilde{x}_s};\\
			&\ \ \ \ \ \ y^s_{k+1} = y^s_k - \eta\tau\cdot \widetilde{\nabla}^{(2)}_{k};\\
			&\tilde{x}_{s+1} = \frac{1}{m}\sum_{k=1}^m {y^s_{k+1}}; \\
			&y^{s+1}_1 = y^s_{m+1} + \underline{(1 - \tau) (\tilde{x}_{s+1} - \tilde{x}_s)};
		\end{aligned}
		\]
	\end{minipage}
\end{center}

The underlined parts of both algorithms can be regarded as the source of acceleration, since setting $\tau = 1$ makes both algorithms degenerate into SAGA or Prox-SVRG\footnote{In fact, setting $\tau = 1$ does not make SSNM and MiG exactly the same as SAGA and Prox-SVRG. For SSNM, the update index for the ``points'' table is different; for MiG, the initial point $y^{s+1}_1$ for the new epoch is different.}. A more careful analysis shows that: For MiG, the momentum $\tilde{x}_{s+1} - \tilde{x}_s$ is provided every $m$ stochastic steps, where $m = \Theta(n)$ as suggested by the analysis in~\citep{kw:MiG}; for SSNM, although a little bit messy in randomness, we can observe that in expectation, every $n$ steps, the momentum is provided by the newly computed iterate. In comparison, the momentum in Acc-Prox-SVRG~\citep{nitanda:svrg} is added in every stochastic step. However, as analyzed in~\citep{nitanda:svrg}, in pure stochastic setting (mini-batch size is 1)\footnote{Pure stochastic setting is important since it is proven that in order to achieve the optimal convergence rate per data access, we should always choose a mini-batch size of 1 for a family of variance reduction methods~\citep{stoc_batch}.}, no acceleration can be guaranteed for Acc-Prox-SVRG in theory. The intuition here is that we may not trust the momentum provided in every stochastic step; instead, we trust the momentum provided by the average information of $n$ stochastic steps.

Based on the above observation, we may understand the negative momentum in SSNM and MiG as the Nesterov's momentum based on average information, in addition to attaining tighter variance bounds. 

\section{Experiments}
\label{sec:exp}

In this section, we conducted experiments to examine the practical performance of SSNM as well as to justify our theoretical results. All the algorithms were implemented in
C++ and executed through a MATLAB interface for a fair
comparison. We ran experiments on an HP Z440 machine with a single Intel Xeon E5-1630v4 with 3.70GHz cores, 16GB RAM, Ubuntu 16.04 LTS with GCC 4.9.0, MATLAB R2017b.

We are optimizing the following binary problem with $a_i \in \R^d$, $b_i \in \lbrace -1, +1\rbrace$, $i=1\ldots m$:
\[
\text{Logistic Regression:  }\ \  \frac{1}{n} \sum_{i=1}^n {\log{(1 + \exp{(-b_ia_i^Tx))}}} + \frac{\lambda}{2} \norm{x}^2,
\]
where $\lambda$ is the regularization parameter and all the datasets used were normalized before running the experiments.

The experiments were designed as some ill-conditioned problems (with very small $\lambda$), since ill-condition is where all the accelerated first-order methods take effect. We tested the following algorithms with their corresponding parameter settings:

\begin{itemize}
	\item SAGA. We set the learning rate as $\frac{1}{2(\mu n + L)}$, which is analyzed theoretically in~\citep{defazio:saga}.
	\item SSNM. We used the same settings as suggested in Algorithm~\ref{SIAM_SC}, which are $\eta = \sqrt{\frac{1}{3\mu n L}}$ and $\tau = \frac{n\eta \mu}{1 + \eta\mu}$.
	\item Katyusha. As suggested by the author, we fixed $\tau_2 = \frac{1}{2}$, set $\eta = \frac{1}{3\tau_1 L}$ and chose $\tau_1 = \sqrt{\frac{m}{3\kappa}}$~\citep{zhu:Katyusha} (In the notations of the original work).
	\item MiG. We set $\eta = \frac{1}{3 \theta L}$ and chose $\theta = \sqrt{\frac{m}{3\kappa}}$ as analyzed in~\citep{kw:MiG}.
\end{itemize}
\begin{figure*}[t]
	\begin{center}
		\centerline{
			\includegraphics[width=0.26\textwidth]{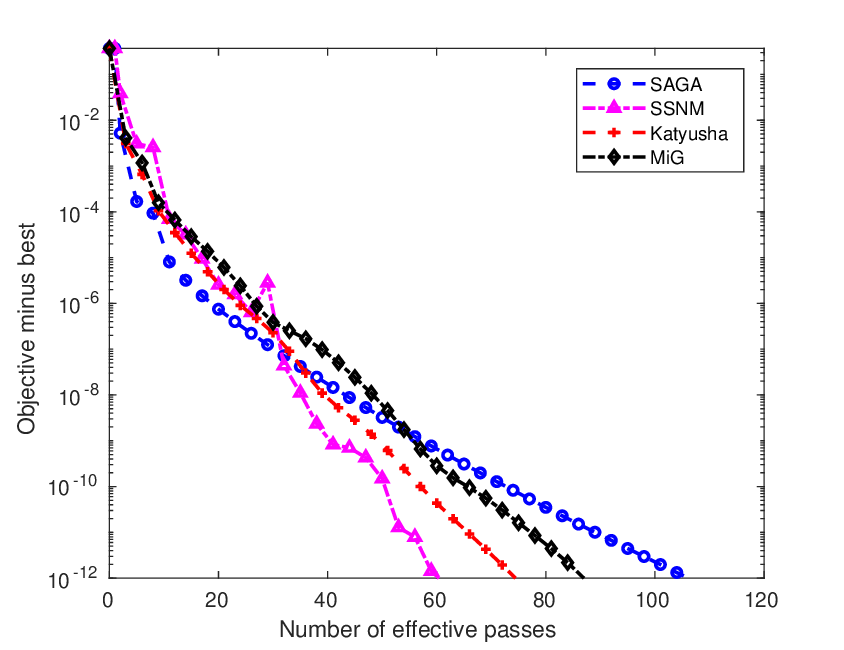}
			\includegraphics[width=0.26\textwidth]{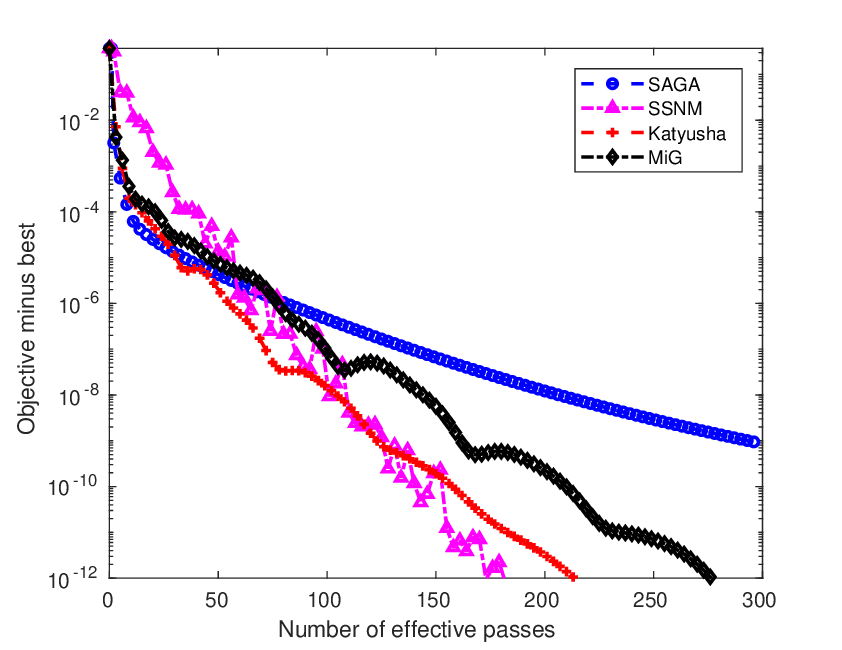}
			\includegraphics[width=0.26\textwidth]{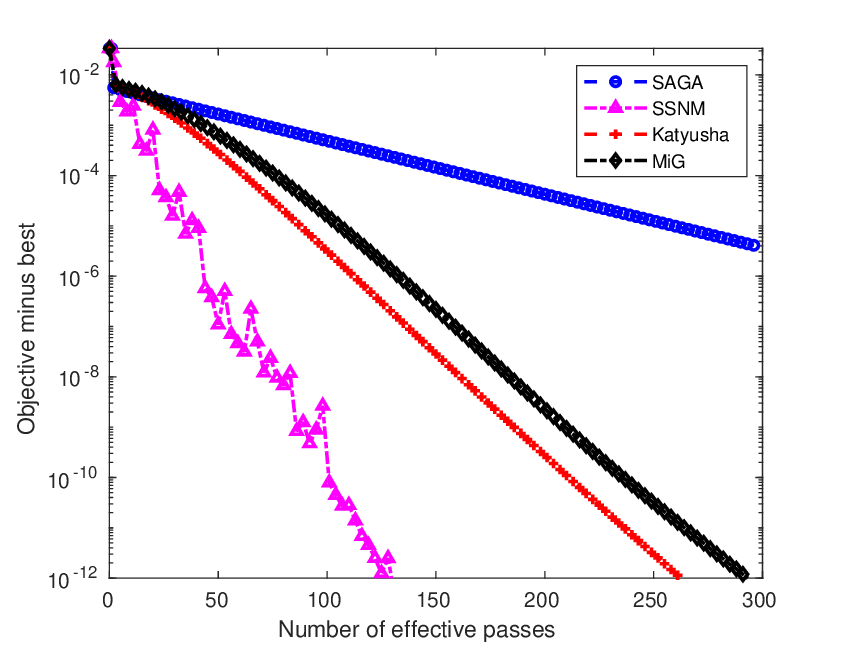}
			\includegraphics[width=0.26\textwidth]{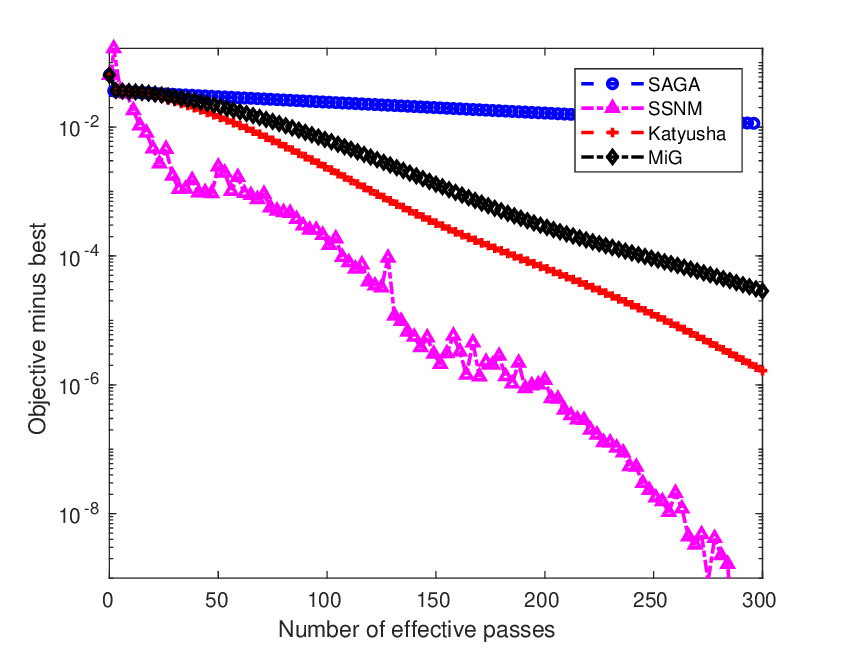}
		}
		\caption{Evaluations of SAGA, SSNM, Katyusha and MiG on the \textsf{a9a} dataset with $\lambda = 10^{-6}$ and $10^{-7}$ (the first two figures) and the \textsf{covtype} dataset with $\lambda = 10^{-8}$ and $10^{-9}$ (the last two figures).}
		\label{exp_practice_1}
	\end{center}
\end{figure*}

We report the results in Figure~\ref{exp_practice_1}. From the results, we can make the following observations to justify the accelerated convergence rate:
\begin{itemize}
	\item \textit{Similar convergence results comparing with other accelerated algorithms.} In fact, we are surprised by the excellent performance of SSNM on the \textsf{covtype} dataset. For this dataset, SSNM is even significantly faster than Katyusha and MiG in terms of the number of epochs (though in theory, Katyusha and MiG yield the same convergence rate as SSNM). The fast convergence of SSNM in practice imply that the algorithm could potentially benefit many applications.
	\item \textit{Around 3 times slow-down when $\kappa$ is 10 times larger.} It can be observed that using the same dataset, when we divide $\lambda$ by 10 (the same as multiply $\kappa$ by 10), approximately $\sqrt{10}$ times slow-down ($\sqrt{10}$ times more oracle calls required to achieve the same accuracy) is recorded for all the accelerated methods. In comparison, SAGA shows significant slow-down when $\kappa$ is increased in both experiments. This observation justifies the $\sqrt{\kappa}$ dependency for accelerated methods.
\end{itemize}

Another observation is that accelerated methods seem to perform worse in the experiments on the \texttt{a9a} dataset at first several passes. We conjecture that this is because the objective is locally well-conditioned around the initial point. For well-conditioned problem, accelerated methods do not yield a faster rate in theory. In practice, we always found that a smaller amount of momentum yields a better performance. Non-accelerated methods (SVRG, SAGA) always perform better in this case, since they are the accelerated methods without momentum. In the parameter schemes of SSNM, MiG, and Katyusha, the amounts of negative momentum are all set to be $\geq 1/2$ for simplicity in the proofs. To achieve more consistent performance, we can derive parameter schemes that have a smaller amount of momentum.

However, as also reported in Figure~\ref{exp_practice_1}, the convergence of SSNM, though very fast, is somewhat unstable compared with the other three methods. This can be explained by the double sampling trick used in SSNM, which greatly increases the uncertainty inside each iteration.

An empirical comparison with Point-SAGA for ridge regression is also provided in Appendix~\ref{sec:memo_point} for reference.

\subsection{Effectiveness of sample $I_k$}

\begin{wrapfigure}{r}{0.37\textwidth}
	\vspace{-40pt}
	\begin{center}
		\includegraphics[width=0.32\textwidth]{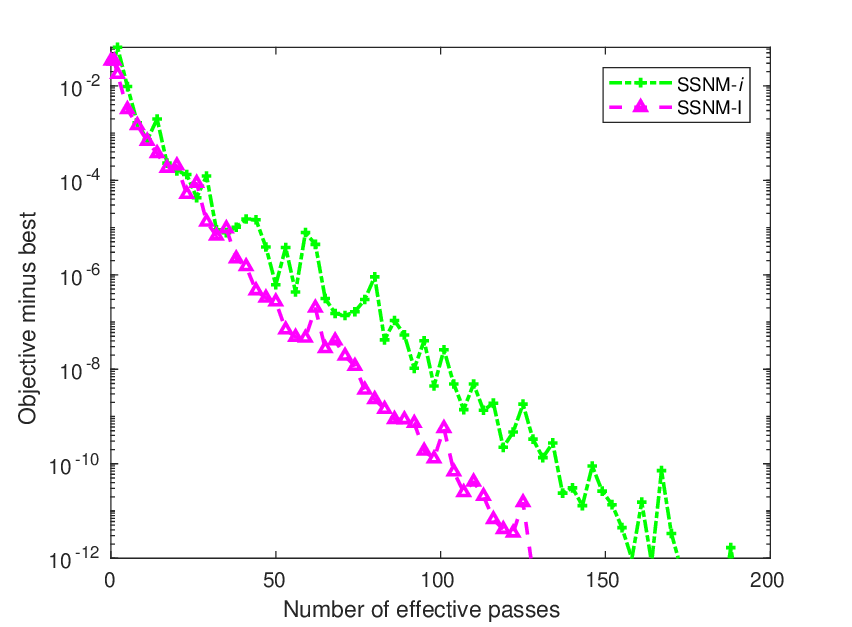}
		\caption{Comparison of using sample $i_k$ (SSNM-\textit{i}) or $I_k$ (SSNM-I) in $7$th step of SSNM on \texttt{covtype} with $\lambda = 10^{-8}$.}
		\label{exp_practice_2}
	\end{center}
	\vspace{-40pt}
\end{wrapfigure}

A natural question is that: can we use sample $i_k$ (the sample of stochastic gradient) instead of an independent sample $I_k$ in the $7$th step of Algorithm~\ref{SIAM_SC}? We empirically evaluated the effect of sample $I_k$ as shown in Figure~\ref{exp_practice_2}. As we can see, using sample $i_k$ makes the algorithm even more unstable and slower in convergence comparing with using an independent sample $I_k$. This effect can probably be explained by some kind of variance cumulation when using the sample $i_k$.

\section{Conclusions}

In this paper, we proposed SSNM, an accelerated variant of SAGA, which uses the \textit{Sampled Negative Momentum} trick. Our theoretical results show that SSNM achieves the best known oracle complexity for strongly convex problems and our experiments justified such improvements for the ill-conditioned problems. Regarding its superiority over SAGA and Point-SAGA in convergence rate or oracle requirement, SSNM is potentially beneficial for a large family of high-dimensional machine learning tasks.

\bibliographystyle{abbrvnat}
\bibliography{SIAM}
\newpage
\appendix
\section{An empirical comparison with Point-SAGA}
\label{sec:memo_point}

\begin{wrapfigure}{r}{0.37\textwidth}
	\vspace{-27pt}
	\begin{center}
		\includegraphics[width=0.4\textwidth]{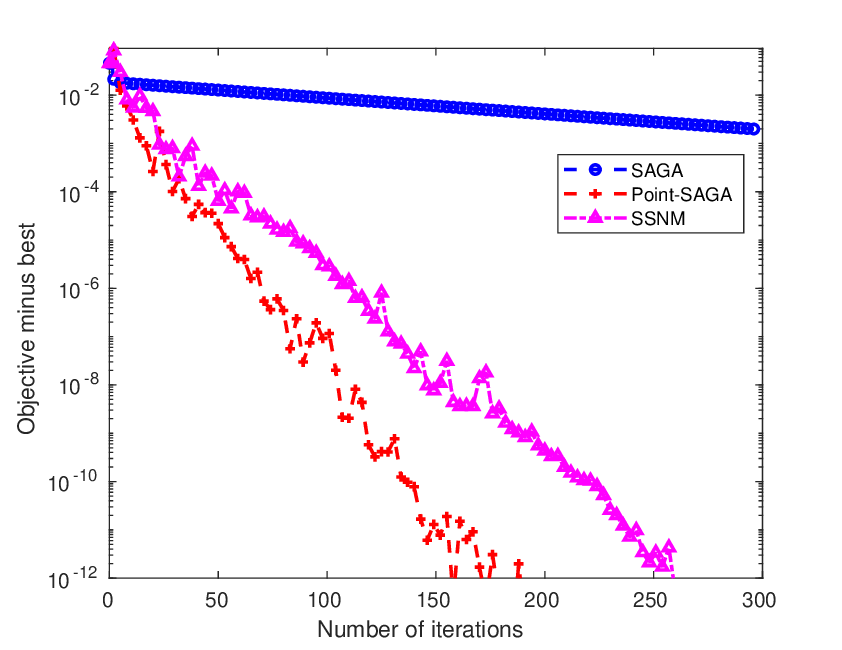}
		\vspace{-15pt}
		\caption{Comparison of SAGA, Point-SAGA and SSNM for solving ridge regression on \texttt{covtype} with $\lambda = 10^{-8}$.}
		\label{exp_point}
	\end{center}
	\vspace{-25pt}
\end{wrapfigure}

Here we report an experiment comparing the performance of SAGA, Point-SAGA and SSNM with respect to iteration counter. The detailed experimental setting is given in Section~\ref{sec:exp} in the main paper. Since Point-SAGA requires the exact proximal operator of each $F_i(\cdot)$ in theory, we focus on training ridge regression in this section:
\[
\textit{Ridge Regression: } \frac{1}{n} \sum_{i=1}^n {\frac{1}{2}(a_i^Tx + b_i)^2} + \frac{\lambda}{2} \norm{x}^2.
\]
Note that the proximal operator of each $F_i(\cdot) = \frac{1}{2}(a_i^Tx + b_i)^2 + \frac{\lambda}{2} \norm{x}^2$ can be efficiently computed as mentioned in~\citep{defazio:sagab}.

\textbf{A memory issue of Point-SAGA:} In fact, when we involve an $\ell 2$-regularizer in each $F_i(\cdot)$ \footnote{An $\ell 2$-regularizer is always the source of strong convexity for real world problems.}, we cannot use the trick of representing a gradient by a scalar since the update equation of the new table entry $g^{k+1}_j$ (in original notations) contains terms that correlate to the weight $x_k$ and the running average, which leads to an $O(nd)$ memory complexity. A possible solution is to separate the proximal operator computations for the component functions and the regularizer, but it does not fit in the analysis of Point-SAGA. 

We used the same parameter settings for SAGA and SSNM as in Section~\ref{sec:exp} in the main paper. For Point-SAGA, we chose the learning rate $\gamma$ suggested by the original work~\citep{defazio:sagab},
\[
\gamma = \frac{\sqrt{(n - 1)^2 + 4n\frac{L}{\mu}}}{2Ln} - \frac{1 - \frac{1}{n}}{2L}.
\]

The result is shown in Figure~\ref{exp_point}. As we can see, the convergence rates of Point-SAGA and SSNM are quite similar and consistently faster than SAGA. Although Point-SAGA is shown to be slightly faster than SSNM in this experiment, considering the general objective assumption and the memory issue of Point-SAGA mentioned above, SSNM is a more favorable accelerated variant of SAGA than Point-SAGA in practice. Interestingly, both accelerated variants are more unstable than SAGA in this experiment. 

\end{document}